\documentclass[10pt,twocolumn,letterpaper]{article}

\usepackage{cvpr}              % To produce the CAMERA-READY version
\usepackage{array,tabularx}
\newcolumntype{L}{>{\raggedright\arraybackslash}X}
\usepackage{multirow}
\usepackage{diagbox}
\usepackage[accsupp]{axessibility}

\definecolor{cvprblue}{rgb}{0.21,0.49,0.74}
\usepackage[pagebackref,breaklinks,colorlinks,allcolors=cvprblue]{hyperref}

\def\paperID{19814} % *** Enter the Paper ID here
\def\confName{CVPR}
\def\confYear{2026}

\newcommand{\dataset}{WebChain}

\title{\dataset: A Large-Scale Human-Annotated Dataset of Real-World Web Interaction Traces}
\author{
Sicheng Fan$^{1,2,3}$ \quad
Rui Wan$^{1}$\quad
Yifei Leng$^{2,3}$ \quad
Gaoning Liang$^{2,3}$ \\
Li Ling$^{1}$ \quad Yanyi Shang$^{2,3}$ \quad Dehan Kong$^{2,3}$ \\
$^{1}$Fudan University \quad $^{2}$IMean AI \quad $^{3}$WebAgentLab
}

\hypersetup{
  colorlinks=true,
  linkcolor=MidnightBlue,
  citecolor=MidnightBlue,
  urlcolor=MidnightBlue
}

\begin{document}
\maketitle
\begin{abstract}
We introduce \textbf{\dataset}, the largest open-source dataset of human-annotated trajectories on real-world websites, designed to accelerate reproducible research in web agents. It contains \textbf{31,725} trajectories and \textbf{318k} steps, featuring a core Triple Alignment of visual, structural, and action data to provide rich, multi-modal supervision. The data is collected via a scalable pipeline that ensures coverage of complex, high-value tasks often missed by synthetic methods. Leveraging this dataset, we propose a Dual Mid-Training recipe that decouples spatial grounding from planning, achieving state-of-the-art performance on our proposed \textbf{WebChainBench} and other public GUI benchmarks. Our work provides the data and insights necessary to build and rigorously evaluate the next generation of scalable web agents. The code and resources are publicly available at \url{https://github.com/franskey-0112/WebChain}.
\end{abstract}

\section{Introduction} \label{sec:intro}

Conquering the browser is one of the most valuable problems in the field of GUI agents, as the web browser serves as the primary interface for the vast majority of digital tasks for human users. Recent advances in Vision-Language-Action (VLA) modeling have sparked hope for agents capable of precise spatial grounding, robust reasoning, and safe long-horizon decision-making\cite{hong2024cogagent,qin2025uitars}. Due to the complex and dynamic nature of web environments and the inherent lack of trajectory data, there is a persistent need for \emph{up-to-date, large-scale, high-quality} datasets in the web agent domain. Current data requirements range from multi-modality spatial grounding, which involves accurately identifying and interacting with UI elements based on visual and textual information, to long-horizon planning trajectories that enable agents to execute complex and multi-step tasks, are becoming increasingly refined with the diversification of evaluation methods and benchmarks.

Existing data efforts face significant limitations. Open-source, human-annotated datasets often lack the necessary scale and completeness\cite{deng2023mind2web,lu2024weblinx,zhou2023webarena} (see Table\ref{tab:dataset_comparison}), which obstructs the verification of model scaling effects. On the other hand, while data synthetic methods \cite{pahuja2025explorer,sun-etal-2025-osgenesis} attempt to harvest real-world interactions at a low cost, they are inherently constrained by security mechanisms. These methods often crumble when facing anti-bot detection, CAPTCHAs, or scenarios requiring personal authentication (e.g., logging into a bank account), rendering them unable to capture the most valuable, high-value user workflows.
Furthermore, a significant body of work on scaling GUI agent models relies on proprietary datasets\cite{qin2025uitars,pahuja2025explorer,wang2025uitars2}, derived from either manual annotation or synthetic methods. This practice renders their key insights non-reproducible, which severely impedes the establishment of community consensus and the overall advancement of the field.

\begin{table*}[t]
\centering
\caption{Comparison with Existing Web Interaction Trajectories Datasets}
\label{tab:dataset_comparison}
\resizebox{\textwidth}{!}{%
\begin{tabular}{lccccc}
\toprule
\textbf{Feature} & \textbf{\dataset} & \textbf{Mind2Web} & \textbf{WebArena (Env.)} & \textbf{WebLINX} & \textbf{GUIAct(multi)} \\
\midrule
\multicolumn{6}{l}{\textit{Scale}} \\
Trajectories & \textbf{31725} & 2350 & N/A & 2337 & 5696 \\
Steps & \textbf{318k} & 17155  & N/A & 100k+  & 44k  \\
Websites & \textbf{428} & 137 & 4 Domains & 155 & 121 \\
\midrule
\multicolumn{6}{l}{\textit{Data Source}} \\
Real Websites & \checkmark & \checkmark & \texttimes & \checkmark & \checkmark \\
Human Trajectories & \checkmark & \checkmark & \texttimes & \checkmark & \checkmark \\
\midrule
\multicolumn{6}{l}{\textit{Visual Data}} \\
Viewport Screenshot & \checkmark & \checkmark & \checkmark & \checkmark & \checkmark \\
\midrule
\multicolumn{6}{l}{\textit{Localization}} \\
Pixel-level Coordinates & \textbf{\checkmark} & \checkmark & \checkmark & \checkmark & \checkmark \\
Bounding Boxes & \textbf{\checkmark} & \checkmark & \texttimes & \checkmark & \checkmark \\
Accessibility Tree & \textbf{\checkmark} & \texttimes & \checkmark & \texttimes & \texttimes \\
\midrule
\multicolumn{6}{l}{\textit{Temporal Granularity}} \\
Event Timestamps & \checkmark & \checkmark & \texttimes & \checkmark & \texttimes \\
\bottomrule
\end{tabular}%
}
\end{table*}

To democratize research in this field and break the data monopoly, we introduce \textbf{\dataset}, a fully open-source ecosystem comprising the largest scale web interaction trajectories collected from actual websites with human annotations. \dataset~is totally constructed by human annotators operating on live, diverse websites, ensuring the inclusion of complex tasks that require authentication and nuanced decision-making. Central to our contribution is the \emph{Triple Alignment} mechanism. Our logging pipeline rigorously synchronizes three layers of context: (1) \textbf{Visual Context} via viewport and full-page screenshots; (2) \textbf{Structural Context} via Accessibility (AX) trees; and (3) \textbf{Action Alignment} via precise pixel coordinates, bounding boxes, and CSS selectors. This multi-layered supervision allows models to not only "see" the page but to understand the structural logic behind every pixel, providing rich signals for grounding, attention allocation, and DOM-aware navigation.

Beyond the data itself, we demonstrate that \dataset~serves as a fundamental building block for a new generation of web agents. We introduce a rigorous evaluation suite, \textbf{WebChainBench}, to measure progress in both spatial grounding and long-horizon planning. Through extensive experiments, we reveal critical scaling laws, showing that model performance on long-horizon tasks correlates positively with the scale of our human-verified data. Furthermore, we identify a superior training paradigm—\emph{Dual Mid-Training}—which decouples spatial perception from temporal planning, achieving state-of-the-art (SOTA) performance on both \textbf{WebChainBench} and well-adopted public benchmarks.

In summary, our contributions are three-fold:
\begin{itemize}
    \item \textbf{Largest Human-Annotated Real-World Web Trajectory Dataset:} We release the largest human-annotated dataset collected from real-world websites to date. We make the entire pipeline—including the data, collection tools, and benchmarks—publicly available to foster community innovation.
    
    \item \textbf{Scalable Construction Pipeline:} We propose a robust pipeline combining constraint-based task synthesis, human-in-the-loop verification, and automated context enrichment (including CoT and visual grounding densification), enabling the efficient production of high-quality data at scale.
    
    \item \textbf{Investigation on Training Recipes \& SOTA Performance:} We validate the dataset's value through extensive experiments to identify the optimal training recipe for scaling GUI agent models. Our findings reveal that when building upon a reinforcement finetuning foundation, a \textit{Dual Mid-Training} paradigm—which disentangles spatial grounding from planning—significantly boosts performance. This approach achieves new state-of-the-art results on complex, long-horizon web tasks, outperforming standard supervised methods.
\end{itemize}

\begin{figure*}[t]
  \centering
  \includegraphics[width=0.8\textwidth]{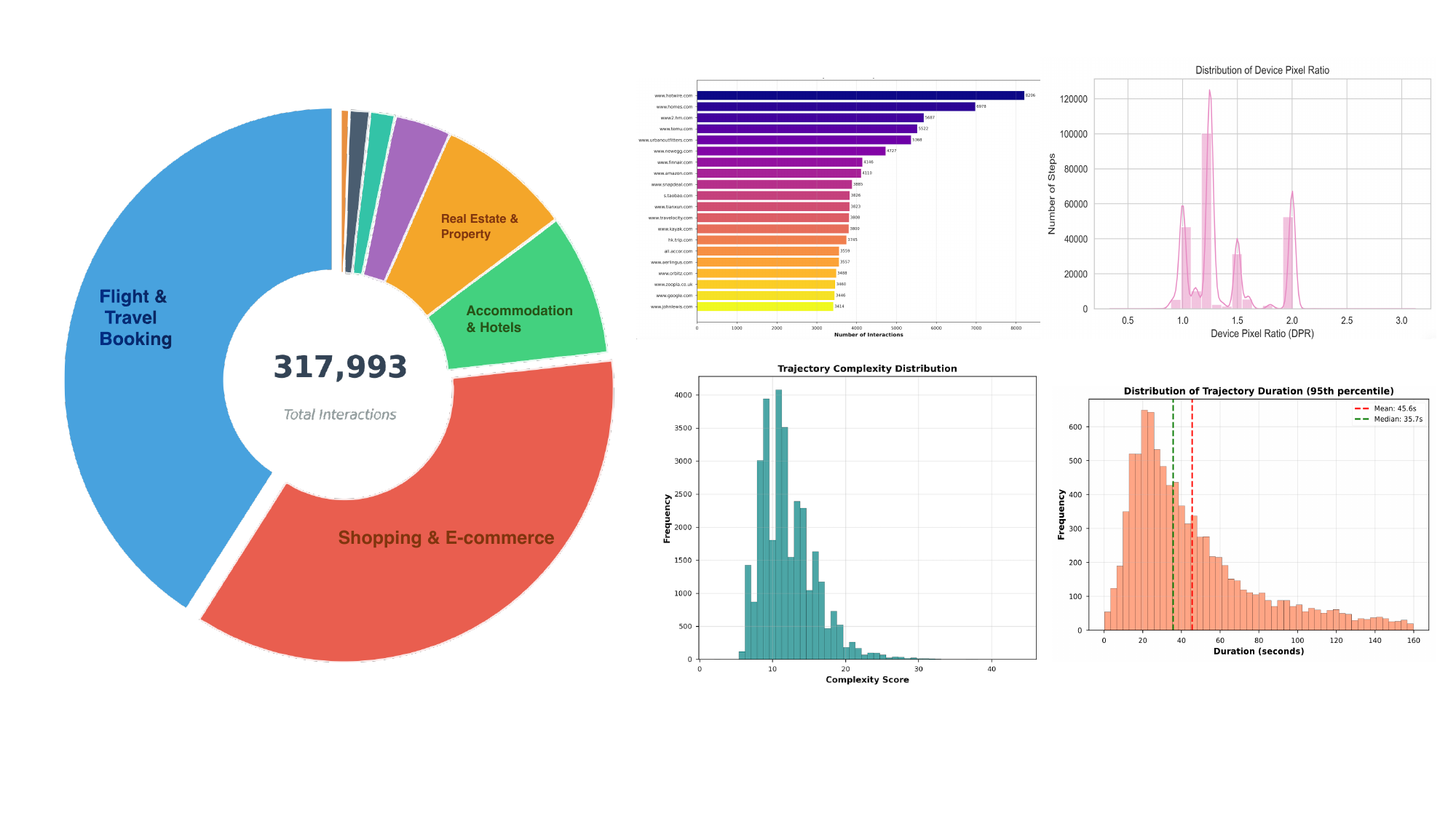}
  \caption{\textbf{Dataset Overview.}
  A statistical summary of \dataset, including interaction distribution across website categories, top-domain interaction frequencies, device pixel ratio distribution, trajectory complexity, and trajectory duration. These statistics collectively highlight the scale and diversity of \dataset.}
  \label{fig:dataset_overview}
\end{figure*}

\section{Related Work}

\subsection{Web Interaction Datasets and Benchmarks}
The evolution of web agents is fundamentally constrained by the availability of training and evaluation environments. Early research relied on synthetic, sandboxed environments like MiniWoB++~\cite{liu2018reinforcement}, which offered controlled tasks but failed to reflect the visual and structural noise of the open web. To bridge this gap, data-driven approaches emerged, mining interactions from real-world applications. Rico~\cite{deka2017rico} pioneered this in mobile UIs, while Mind2Web~\cite{deng2023mind2web} and WebLINX~\cite{lu2024weblinx} extended it to the web, providing large-scale, offline trajectories. However, these datasets still lack the scale required to systematically validate the scaling effects of modern GUI agent models.

To enable interactive evaluation, recent works have introduced realistic execution environments. WebArena~\cite{zhou2023webarena} and VisualWebArena~\cite{koh2024visualwebarena} host reproducible websites but suffer from the "simulation gap"—lacking the evolving DOM structures, ads, and visual clutter of the live web. Conversely, synthetic methods like Explorer~\cite{pahuja2025explorer} and OS-Genesis~\cite{sun-etal-2025-osgenesis} operate on the live web but are inherently limited by security boundaries; they cannot capture high-value workflows requiring authentication (e.g., banking, e-commerce checkout) due to bot detection and privacy constraints. 
\textbf{\dataset} fills this critical void by providing the largest, human-annotated corpus collected on live, diverse websites. Unlike synthetic methods, our human-in-the-loop pipeline captures complex, authenticated trajectories, while our \textit{Triple Alignment} ensures distinct supervision for visual, structural, and action grounding.

\subsection{Vision-Language Models for GUI Grounding}
The paradigm for GUI agents has shifted from DOM-based language models to pixel-aware Vision-Language-Action (VLA) models. Early approaches relied on parsing HTML text~\cite{gur2018learning}, often struggling with the token context limits of complex web pages. Recent advancements, driven by general-purpose VLMs (e.g., Qwen2.5-VL~\cite{bai2023qwen}) and specialized UI models (e.g., SeeAct~\cite{zheng2024gpt}, CogAgent~\cite{hong2024cogagent}), utilize screenshots to perceive layout and visual cues. Techniques such as Set-of-Marks (SoM)~\cite{yang2023set} have been proposed to aid visual grounding by overlaying numeric tags on interactive elements.
Despite these advances, VLMs continue to suffer from spatial hallucinations and grounding errors, particularly on dense, clutter-heavy web pages~\cite{cheng2024seeclick}. This limitation is largely attributed to the scarcity of fine-grained grounding data that aligns high-level intent with precise bounding boxes and DOM elements. Our work addresses this by providing dense, verified spatial annotations, enabling models to decouple perception from planning as demonstrated in our Dual Mid-Training experiments.

\begin{figure*}[t]
  \centering
  \includegraphics[width=\textwidth]{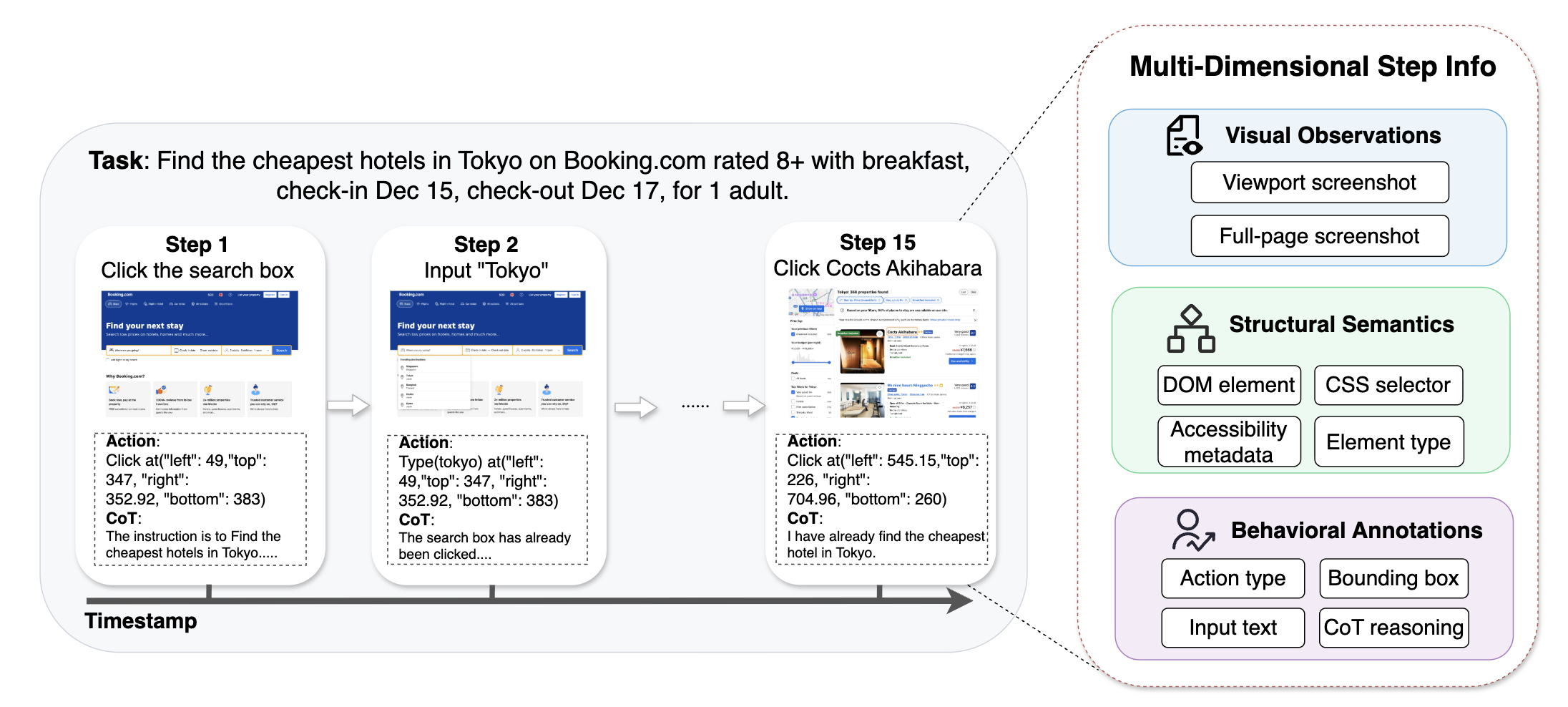}
  \caption{\textbf{Example trajectory and multi-dimensional step information in \dataset.}
  Left: a long-horizon task on Booking.com with key steps along the trajectory.
  Right: the multi-dimensional step schema, including visual observations, structural semantics, and behavioral annotations.}
  \label{fig:step_info}
\end{figure*}

\subsection{Training Paradigms: From SFT to RL}
Supervised Fine-Tuning (SFT) remains the dominant strategy for training web agents, teaching models to mimic expert behavior~\cite{deng2023mind2web}. However, SFT suffers from the distribution shift problem (covariate shift), where agents fail to recover from states deviating from the training trajectory. Reinforcement Learning (RL) offers a compelling alternative by optimizing for long-horizon task success rather than step-by-step imitation. 
Prior works have applied RL to web navigation with promising results in simplified environments~\cite{ouyang2022training, humphreys2022data}. However, scaling RL to modern VLMs on open-ended web tasks introduces significant challenges, primarily due to sparse reward signals and the immense exploration space of the internet. While recent system-level innovations like AReaL~\cite{fu2025areal} and GUI-R1~\cite{luo2025gui} address computational efficiency, the algorithmic stability of RL relies heavily on a "warm start"—a robust initial policy. Our experiments demonstrate that \dataset~serves as this critical foundation. By enabling a high-quality SFT initialization (specifically via our CoT-augmented mid-training), we unlock the stability required for effective RL fine-tuning, achieving state-of-the-art performance on long-horizon benchmarks.

\section{The \dataset~Dataset} \label{sec:dataset}

\subsection{Overview} \label{sub:dataset_overview}
We present \dataset, a comprehensive, large-scale multimodal dataset designed to empower web agents with grounded, long-horizon capabilities. To the best of our knowledge, \dataset~represents the largest corpus of its kind constructed to date. An overview of dataset statistics is shown in Figure~\ref{fig:dataset_overview}. Our dataset distinguishes itself through four core pillars:
(1) \textbf{Unprecedented Scale and Coverage}: We encompass a vast diversity of real-world websites, significantly expanding the task boundaries of existing benchmarks. Notably, we employ an innovative task proposal framework to ensure both the complexity and diversity of tasks across these domains;
(2) \textbf{Rich Multimodality}: We align pixel-space vision, hierarchical DOM-tree structures, and accessibility metadata to provide a holistic view of the web environment;
(3) \textbf{Fully Human-Annotated Trajectories}: Unlike semi-synthetic approaches, our dataset is constructed entirely through rigorous human annotation, ensuring gold-standard reliability and precise alignment between user intent and executable actions;
(4) \textbf{Open Availability}: We fully open-source the \dataset~dataset to the research community, facilitating reproducible research and accelerating advancements in generalist web agents.

\subsection{Data Construction Pipeline} \label{sub:construction_pipeline}

To construct a dataset that ensures both scalability in task diversity and high-fidelity alignment with real-world website interactions, we designed and implemented a three-stage pipeline: (1) \textit{Constraint-Based Task Synthesis}, (2) \textit{Human-in-the-Loop Trajectory Collection}, and (3) \textit{Post-processing Contextual Enrichment}.

\subsubsection{Stage 1: Constraint-Based Task Synthesis}
The primary challenge in task generation is ensuring that synthesized goals are both executable (i.e., not hallucinated) and representative of diverse user intents. Standard LLMs lack the real-time, instance-specific knowledge of a given website's functionalities (e.g., attempting to "sort by user reviews" on a site lacking that feature). We address this via a structured two-step process:

\paragraph{Structured Functionality Extraction.}
Before task generation, we first perform static analysis on each target website to extract a structured functional schema. This schema defines the precise executability boundaries for any potential task. The extraction process parses two layers of information:
\begin{itemize}
    \item \textbf{Domain Semantics:} This layer identifies the site's high-level purpose (e.g., Online Travel Agency, E-commerce, Fintech) and the specific granularity of its services (e.g., "Domestic Flights Only" vs. "International Flights," available product categories).
    \item \textbf{Interactivity \& Logic:} This layer maps the specific interactive components and their underlying logic. This includes enumerating refinement mechanisms such as sorting toggles (e.g., \texttt{sort\_by: price\_asc}), faceted filters (e.g., \texttt{filter: \{brand: [A, B, C], price\_range: [min, max]\}
    }), and crucial conditional dependencies (e.g., a "Select Model" dropdown being populated only after a "Select Brand" action).
\end{itemize}
This schema functions as a formal grounding constraint for the subsequent generation step, ensuring all synthesized tasks are functionally viable and plausible.

\paragraph{Schema-Constrained Task Generation.}
We employ a generator LLM, explicitly conditioned on the extracted functional schema, to synthesize a large corpus of tasks. This approach prevents the generator from hallucinating actions or constraints unsupported by the target website. The tasks are automatically stratified by complexity to ensure comprehensive coverage:
\begin{itemize}
    \item \textbf{Simple Information Retrieval:} Single-step queries for static information verifiable against the extracted metadata (e.g., \textit{"Find the customer service phone number on Example.com"}).
    \item \textbf{Multi-Constraint Navigation:} Goals requiring the logical composition of multiple filters or refinement steps (e.g., \textit{"On Amazon, find a 4K TCL television released in 2023, priced under \$300, and sort the results by popularity."}).
    \item \textbf{Conditional Dependency Tasks:} Goals that require navigating sequential logic, where actions are contingent on previous states (e.g., \textit{"Select a flight from SFO to JFK, and then add priority boarding for the first leg of the journey."}).
\end{itemize}

\subsubsection{Stage 2: Human-in-the-Loop Trajectory Collection} The synthesized tasks serve as precise goals for human annotators. To capture ground-truth interaction trajectories, we utilize \textbf{WebChain Builder}. As annotators attempt to complete the assigned tasks, this tool passively and exhaustively captures a multi-modal record of their interaction traces. For each step, the tool records: \begin{itemize} \item The complete pre- and post-action DOM snapshots. \item The specific action executed (e.g., \texttt{click}, \texttt{type}, \texttt{scroll}). \item High-fidelity spatial information, including viewport coordinates and the bounding box of the target element. \item Element-specific metadata, such as the element's XPath, CSS selectors, and inner text. \end{itemize} This process yields a rich, step-by-step dataset of (State, Action, Reward, Next State) tuples, grounded in complex, real-world web environments.

\subsubsection{Stage 3: Post-processing Contextual Enrichment}
To maximize the dataset's utility for training capable agents, the raw trajectories are augmented using two strategies designed to enhance grounding fidelity and agentic reasoning.

\paragraph{Visual Grounding Densification.}
Standard trajectories only label the single element that was interacted with (the positive example). To teach agents a more robust understanding of layout, we enrich the data by parsing the \textit{entire} viewport. For each state (i.e., webpage snapshot), we extract the bounding boxes, element types (e.g., \texttt{button}, \texttt{input}, \texttt{a}), and text content for \textit{all} interactive elements visible on the screen. This densification process provides comprehensive negative sampling and transforms the task from simple element-clicking to a dense, layout-aware segmentation problem, enabling the agent to distinguish actionable elements from non-actionable text or decorative regions.

\paragraph{Synthetic Rationale Generation (CoT).}
To explicitly model the planning and reasoning required for agentic behavior, we synthesize intermediate reasoning traces (i.e., Chain-of-Thought) for each action. We prompt a powerful Visual Language Model (VLM) with the full trajectory context, including the overarching task goal, the history of (state, action) pairs, and the current visual GUI state. The VLM's task is to generate a natural-language rationale that "thinks aloud," explaining the cognitive process for selecting the next action. For example: \textit{"The goal is to find a \$300 TV. I have already filtered by 'TCL'. Now I see a 'Price Range' filter. I need to click this to enter the price limit."} This exposes the latent cognitive process, providing a supervisory signal that encourages the agent to learn interpretable, multi-step planning.

\subsection{Dataset Statistics and Modalities} \label{sub:dataset_stats}

\dataset~comprises \textbf{31,725} human-verified trajectories across \textbf{428} diverse domains, totaling over \textbf{318k} interaction steps. The corpus captures significant variability in client environments, spanning multiple operating systems, browsers, and viewport resolutions. As shown in Table~\ref{tab:stats}, the dataset features long-horizon dependencies with an average chain length of 10.02 steps.

\begin{table}[t]
  \centering
  \caption{\textbf{High-level statistics of \dataset.} The dataset emphasizes long-horizon reasoning with diverse domain coverage.}
  \label{tab:stats}
  \resizebox{0.9\linewidth}{!}{
  \begin{tabular}{lrr}
    \toprule
    Metric & Value & Description \\ 
    \midrule
    Total Trajectories & 31,725 & Human-verified sessions \\ 
    Total Steps & 317,993 & Atomic interactions \\
    Unique Domains & 428 & E-commerce, OTA, etc. \\
    Avg. Trajectory Length & 10.02 & Median: 9 \\
    Avg. Duration & 1.07 min & Time per task \\
    \bottomrule
  \end{tabular}
  }
\end{table}

\paragraph{Multimodal Annotations.}
Each step in \dataset~provides a comprehensive snapshot of the agent--environment state, as illustrated in Fig.~\ref{fig:step_info}. The core schema includes:
\begin{itemize}
  \item \textbf{Visual Context:} Full-page screenshots and viewport-specific crops.
  \item \textbf{Structural Context:} HTML and Accessibility (AX) tree snapshots.
  \item \textbf{Precise Grounding:} Element bounding boxes (\texttt{rect}), CSS selectors, and pixel coordinates.
  \item \textbf{Reasoning:} Chain-of-Thought(CoT) traces explaining the rationale behind the action.
\end{itemize}

\subsection{Intended Use and Impact} \dataset~is uniquely positioned to advance web agents in three areas: (1) \textbf{Spatial Grounding}, via pixel-perfect element localization aligning vision with DOM semantics; (2) \textbf{Long-Horizon Planning}, supported by complex branching logic and error recovery scenarios; and (3) \textbf{Standardized Evaluation}, enabling rigorous benchmarking of information retrieval and multi-constraint navigation tasks.

\section{Experiments}
We conduct a comprehensive experimental study to unlock the full potential of \dataset~ and provide a clear roadmap for training powerful web agents. After formalizing our task definitions and benchmark protocols, we investigate the central questions facing the field. We examine the following fundamental hypotheses: Does model performance on complex web tasks scale predictably with the amount of our human-verified data? Given the distinct challenges of spatial grounding and long-horizon planning, can a specialized training strategy outperform standard supervision methods? Do these models generalize, demonstrating state-of-the-art performance beyond our own testbed and general public benchmarks?
% -----------------------------------------------------------------------------
% 4.1 Setup and Definitions
% -----------------------------------------------------------------------------
\subsection{Experiment Formulation}

\subsubsection{Task Formulation}
We design two task formulations—\emph{Spatial Grounding} and \emph{Long-horizon planning}—based on the multi-modal structure of \dataset~and prior web-interaction studies~\cite{luo2025gui}. Both tasks are trained under a unified reward-weighted optimization framework.

\paragraph{Reward Definition.}
At each interaction step $t$, the model receives a scalar reward $r_t \in [0,1]$ composed of two parts:
\[
r_t = \alpha r^{\text{type}}_t + (1-\alpha)r^{\text{content}}_t.
\]
The \emph{action-type reward} $r^{\text{type}}_t$ is $1$ if the predicted action type $\hat{a}_t$ matches the ground-truth type $a_t^\ast$ (e.g., \texttt{click}, \texttt{type}, \texttt{scroll}); otherwise, it is $0$.
The \emph{action-content reward} $r^{\text{content}}_t$ is $1$ if the predicted action arguments $\hat{y}_t$ satisfy the correctness criteria of the corresponding action type. For example:
(i) for a \texttt{click} action, $\hat{y}_t$ must fall inside the ground-truth bounding box $b_t^\ast$;
(ii) for a \texttt{type} action, the predicted text must be a lexical superset of the target string $y_t^\ast$;

\paragraph{Spatial Grounding.}
This task focuses on local spatial alignment. Given a screenshot $I_t$ and a low-level instruction $x_t$, the model $\mathcal{M}$ predicts $(\hat{a}_t, \hat{y}_t)$. The objective is maximizing the expected step reward:
\[
\max_{\theta} \; \mathbb{E}_{(I_t, x_t)} \big[r_t(\hat{a}_t, \hat{y}_t)\big].
\]

\paragraph{Long-horizon planning.}
Long-horizon planning models sequential decision-making. Each instance consists of a global goal $g$, observation $I_t$, and history $h_{t-1}$. The optimization is defined as:
\[
\max_{\theta} \; \mathbb{E}_{(g, I_t, h_{t-1})} \big[r_t(\hat{a}_t, \hat{y}_t)\big].
\]

\begin{table*}[t]
\centering
\small
\setlength{\tabcolsep}{2.4pt}
\renewcommand{\arraystretch}{1}
\begin{tabular}{l|ccc|ccc|ccc|ccc|ccc|ccc|c}
\hline
\multirow{2}{*}{\textbf{Models}} &
\multicolumn{3}{c|}{\textbf{AC-High\cite{li2024effects}}} &
\multicolumn{3}{c|}{\textbf{AC-Low\cite{li2024effects}}} &
\multicolumn{3}{c|}{\textbf{GUI-Act-Web\cite{chen2025guicourse}}} &
\multicolumn{3}{c|}{\textbf{GUI-Odyssey\cite{lu2025guiodyssey}}} &
\multicolumn{3}{c|}{\textbf{OA-Desktop\cite{kapoor2024omniact}}} &
\multicolumn{3}{c|}{\textbf{OA-Web\cite{kapoor2024omniact}}} &
\multirow{2}{*}{\textbf{Overall}} \\ 
\cline{2-19}
 & Type & GR & SR & Type & GR & SR & Type & GR & SR & Type & GR & SR & Type & GR & SR & Type & GR & SR \\ 
\hline
\multicolumn{14}{l}{\textbf{Zero Shot}} \\ 
\hline
Qwen2.5-VL-3B  & 49.1 & 46.1 & 38.7 & 61.0 & 74.6 & 60.0 & 56.5 & 65.9 & 57.3 & 37.9 & 27.1 & 26.9 & 58.1 & 49.0 & 49.0 & 51.2 & 47.0 & 47.0 & 50.1  \\
Qwen2.5-VL-7B  & 68.6 & 62.3 & 57.0 & \textbf{87.0} & 83.0 & 72.0 & 86.5 & 84.2 & 83.9 & 55.1 & 38.1 & 34.7 & 84.0 & 79.3 & 79.3 & 80.3 & 70.3 & 70.3 & 70.9  \\
\hline
\multicolumn{14}{l}{\textbf{Trained on Other Datasets}} \\ 
\hline
GUI-R1-3B    & 58.0 & 56.2 & 46.6 & 83.7 & 81.6 & 64.4 & 89.9 & 87.4 & 76.3 & 54.8 & 41.5 & 41.3 & 91.9 & 78.4 & 78.3 & 88.6 & 75.1 & 75.1 & 70.5  \\
GUI-R1-7B    & 71.6 & 65.6 & 51.7 & 85.2 & 84.0 & 66.5 & 90.9 & 88.1 & 80.3 & 65.5 & 43.6 & 38.8 & 92.2 & 83.4 & 83.3 & 91.2 & 77.3 & 77.3 & 74.2  \\
\hline
\multicolumn{14}{l}{\textbf{Trained on WebChain}} \\ 
\hline
WebChain-LCRL-3B   & 76.9 & 57.1 & 56.7 & 77.9 & 76.2 & 60.8 & 92.3 & 85.6 & 82.6 & 81.4 & 48.7 & 45.7 & 96.9 & 74.2 & 74.2 & 92.6 & 71.3 & 71.3 & 73.5  \\
\textit{+CoT-SFT}      & 79.2 & 59.6 & 57.8 & 80.1 & 79.1 & 64.4 & 94.2 & 88.1 & 84.8 & 84.2 & 50.1 & 46.6 & 98.2 & 75.6 & 75.6 & 93.3 & 72.4 & 72.4 & 75.3  \\
\textit{+SGRL+CoT-SFT}    & 81.1 & 61.0 & 59.4 & 85.2 & 83.1 & 67.2 & 95.3 & 89.6 & 85.5 & 85.5 & 54.4 & 49.2 & 97.8 & 77.5 & 77.5 & 93.7 & 73.8 & 73.8 & 77.3  \\
WebChain-LCRL-7B   & 83.5 & 66.8 & 60.4 & 79.0 & 75.5 & 62.6 & 94.6 & 87.4 & 84.6 & 84.4 & 54.0 & 50.1 & 99.1 & 84.5 & 84.1 & 94.1 & 73.9 & 73.9 & 77.4  \\
\textit{+CoT-SFT}      & \textbf{86.7} & 68.3 & \textbf{62.4} & 81.1 & 77.3 & 63.7 & 95.5 & 88.8 & 86.2 & 86.1 & 56.3 & 52.7 & \textbf{99.7} & \textbf{87.1} & \textbf{86.8} & 95.0 & 74.3 & 74.3 & 79.0  \\
\textit{+SGRL+CoT-SFT}    & 86.2 & \textbf{71.7} & 61.8 & 86.2 & \textbf{84.8} & \textbf{74.1} & \textbf{96.3} & \textbf{90.4} & \textbf{87.6} & \textbf{88.7} & \textbf{57.9} & \textbf{54.8} & 99.5 & 86.2 & 86.2 & \textbf{96.2} & \textbf{78.9} & \textbf{78.9} & \textbf{81.4}  \\
\hline
\end{tabular}
\caption{Performance comparison of models across GUI-Act and OmniAct benchmarks. (\textbf{AC}=\textbf{AndroidControl}, \textbf{OA}=\textbf{OmniAct})}
\label{tab:public-benchmarks}
\end{table*}

\subsubsection{Custom Benchmark: WebChainBench}
To evaluate generalization, we construct \textbf{WebChainBench (WCB)}, consisting of 1.2k interactive steps sampled from held-out \dataset~data. We derive two variants: \textbf{WCB-S} for spatial grounding and \textbf{WCB-L} for long-horizon planning. The dataset is balanced across short ($<6$), medium ($6$--$10$), and long ($>10$ steps) trajectories (Table~\ref{tab:WebChainbench}). Steps are marked correct only when both action type and behavior match.

\begin{table}[h]
\centering
\renewcommand{\arraystretch}{0.9}
\setlength{\tabcolsep}{4pt}
\begin{tabular}{lccc}
\toprule
\textbf{Action} & \textbf{Short ($<6$)} & \textbf{Medium ($6–10$)} & \textbf{Long ($>10$)} \\
\midrule
Click & 362 & 358 & 365 \\
Type & 38 & 42 & 35 \\
\bottomrule
\end{tabular}
\caption{Distribution of WebChainBench samples.}
\label{tab:WebChainbench}
\end{table}

% -----------------------------------------------------------------------------
% 4.2 Data Scaling (Validating the dataset value)
% -----------------------------------------------------------------------------
\subsection{Analysis of Data Scalability}
\label{sec:scaling_law}

To assess the benefits of \dataset's large-scale collection, we examine how the volume of training data influences long-horizon planning performance. Specifically, we perform \textbf{L}ong-\textbf{C}hain-oriented \textbf{RL}VR-based post-training (\textbf{LCRL}) of Qwen2.5-VL-3B on subsets of \dataset~comprising 4k, 20k, and the full set of action steps. All models are trained with identical hyperparameters, using a batch size of 512.

\begin{figure}[htbp]
     \centering
     \includegraphics[width=0.8\columnwidth]{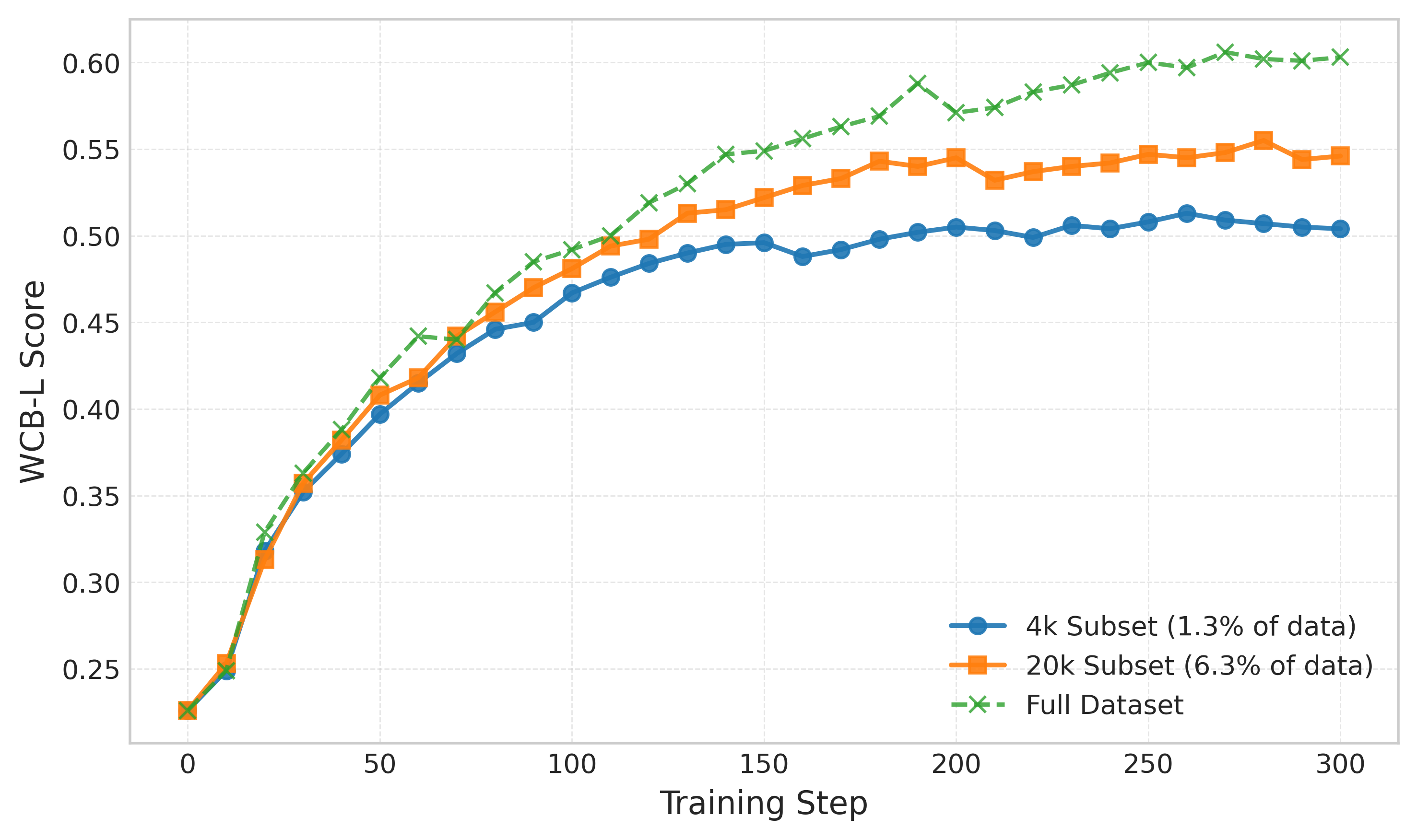} 
     \caption{Scaling effects of $\text{\dataset}$ subsets (4k, 20k, and Full) on Qwen2.5-VL-3B's success rate after LCRL post-training.}
     \label{fig:data_scaling}
 \end{figure}

As illustrated in Figure~\ref{fig:data_scaling}, we observe a clear positive correlation between data volume and model performance on the \textbf{WCB-L} benchmark. The model trained on the full 150k subset achieves significantly higher success rates and follows longer command chains compared to the 4k baseline. This confirms that the scale of \dataset~is instrumental in unlocking robust long-horizon planning capabilities in VLM agents.

% -----------------------------------------------------------------------------
% 4.3 Grounding Optimization (WCB-S)
% -----------------------------------------------------------------------------
\subsection{Training Paradigms for Spatial Grounding}
\label{sec:grounding_opt}

Spatial grounding constitutes a fundamental capability for effective web interaction. In this work, we directly adopt the reward function and task definition introduced in Section~4.1.1 to conduct \textbf{S}patial-\textbf{G}rounding-oriented \textbf{RL}VR training (SGRL). To systematically investigate the optimal training paradigm for this task, we evaluate models on the \textbf{WCB-S} benchmark and hypothesize that two factors are particularly influential:
\begin{enumerate}
\item \textbf{Reasoner Prompting (RP):} Investigates the effect of incorporating prompt-based guidance that elicits explicit, step-wise reasoning about element attributes prior to coordinate prediction.
\item \textbf{Visual Grounding Densification (VGD):} Examines the impact of employing the augmented instruction-action pairs introduced in Section~\ref{sub:construction_pipeline}, which expands the diversity of interactive UI elements (e.g., buttons, input fields).
\end{enumerate}

\begin{figure}[htbp]
     \centering
     \includegraphics[width=0.8\columnwidth]{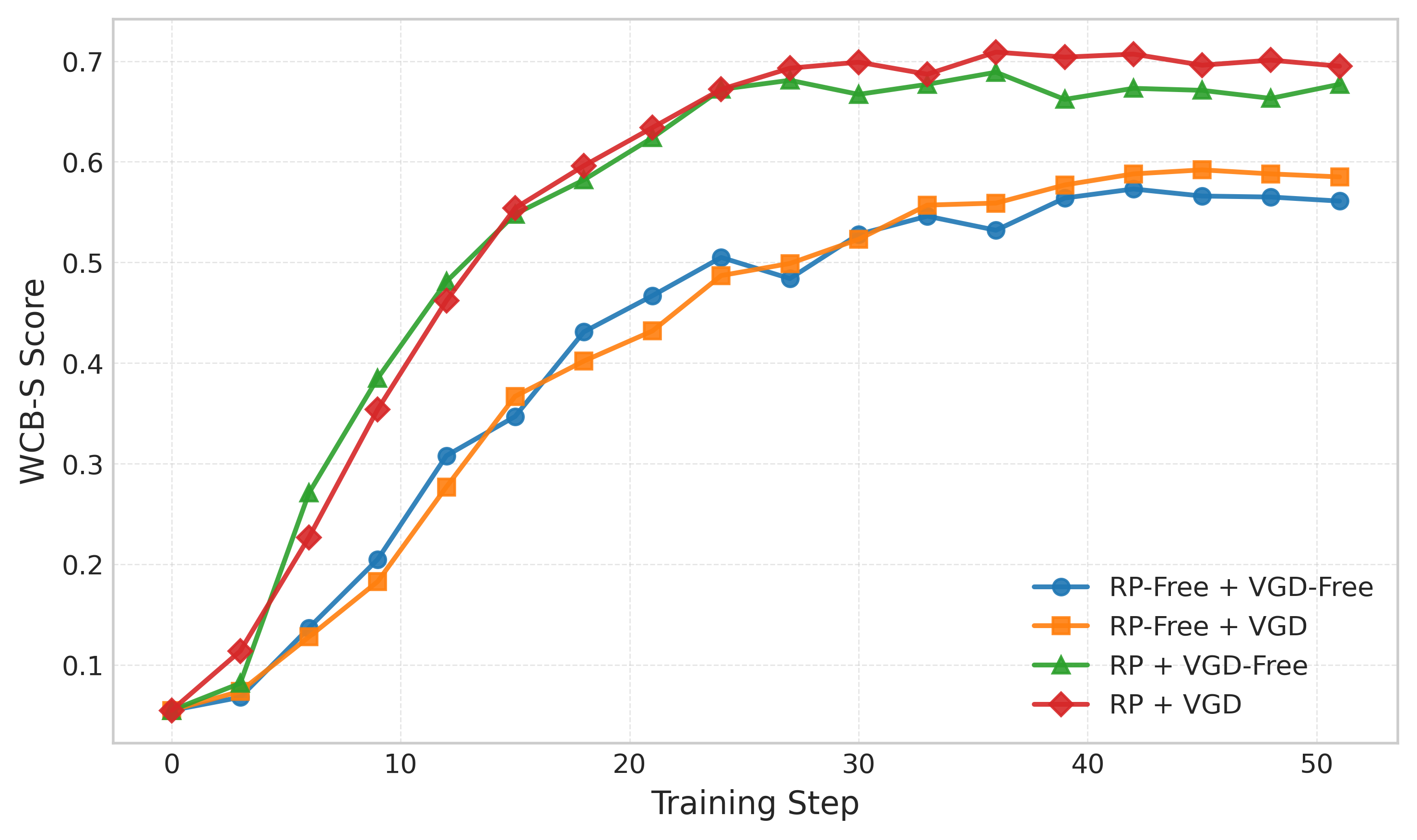} 
     \caption{Study on WCB-S evaluating the effects of Visual Grounding Densification (VGD) and Reasoner Prompting (RP) on spatial grounding performance.}
     \label{fig:grounding_ablation}
 \end{figure}

We evaluate the contribution of these two factors through four controlled experimental settings (\textit{Baseline SGRL training, +VGD, +RP}, and \textit{+Both}). The results, summarized in Table~\ref{fig:grounding_ablation}, reveal that both RP and VGD independently yield measurable gains on WCB-S. Notably, their combination produces the strongest overall performance. Specifically, VGD enhances recall and accurate identification of interactive elements by diversifying training patterns, whereas RP reduces spatial hallucinations through structured reasoning. These results suggest that integrating dataset-level diversity with explicit reasoning supervision constitutes the most effective paradigm for independent spatial grounding task.

% -----------------------------------------------------------------------------
% 4.4 Planning Optimization (WCB-L)
% -----------------------------------------------------------------------------
\subsection{Training Paradigms for Long-horizon Planning}
\label{sec:planning_opt}

Building on grounding capabilities, we investigate high-level planning on the \textbf{WCB-L} benchmark using a multi-stage pipeline: \emph{Mid-Training} followed by \emph{LCRL Post-Training}. Mid-training serves as an intermediary that transfers capabilities and stabilizes learning, leveraging domain-specific, high-quality data to foster structured reasoning and robust spatial grounding, thus laying a strong foundation for subsequent LCRL optimization.

\subsubsection{Impact of Visual Mid-Training Strategies}
\label{sec:visual_impact}

In this section, we investigate how the quality of the underlying visual backbone influences subsequent RL optimization, under the hypothesis that strong spatial grounding is a prerequisite for effective long-horizon control. To test this, we use the four spatial-grounding checkpoints derived from Sec.~\ref{sec:grounding_opt} as initialization and conduct LCRL training on all four models.

\begin{figure}[htbp]
     \centering
     \includegraphics[width=0.8\columnwidth]{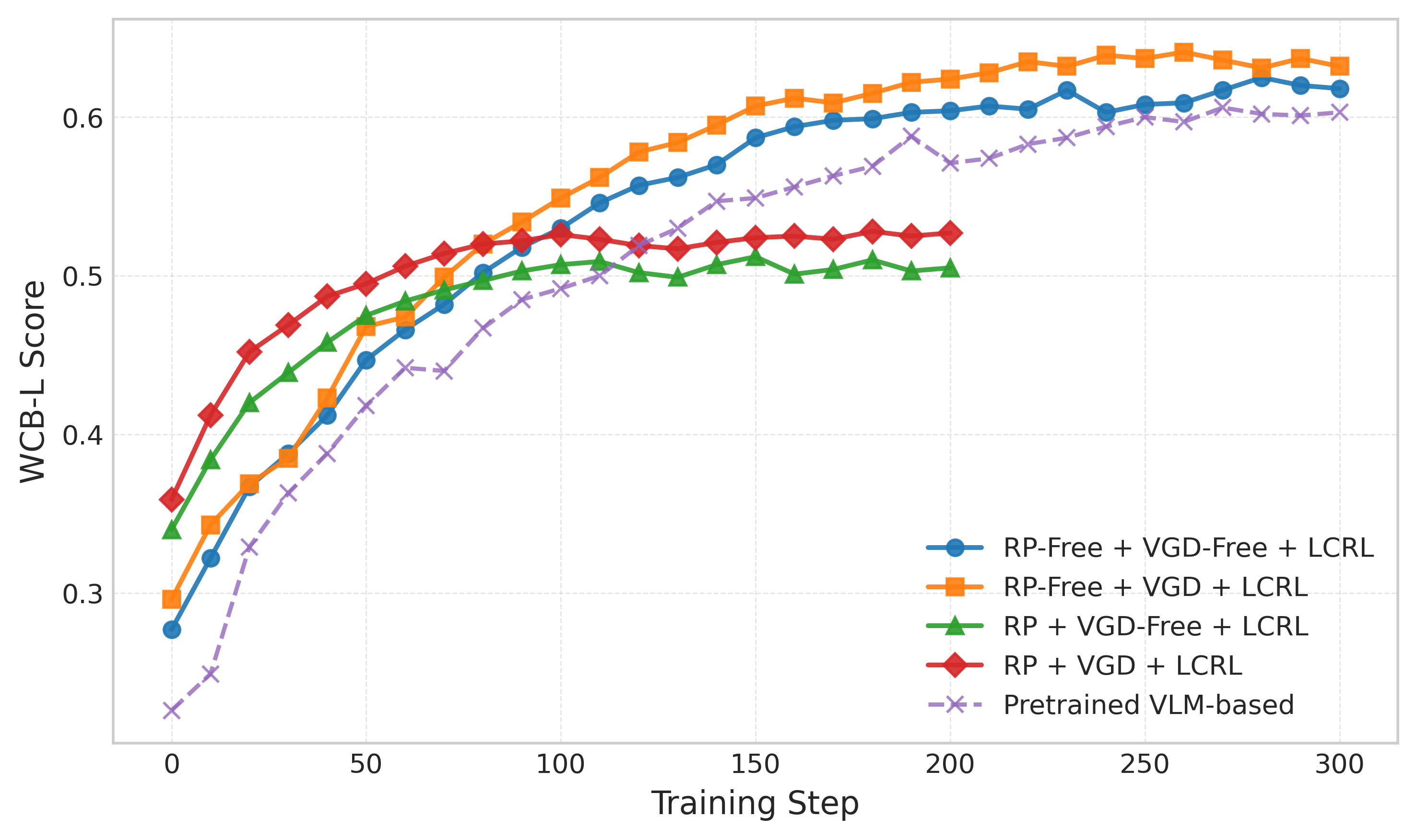} 
     \caption{Study on WCB-L evaluating the effect of SGRL Mid-Training on LCRL Post-Training.}
     \label{fig:visual_mid_impact}
 \end{figure}

\textbf{Analysis}. As shown in Figure~\ref{fig:visual_mid_impact}, the choice of spatial-grounding mid-training strategy has a substantial impact on the ceiling of RL performance. We observe that models initialized from RP-based checkpoints exhibit consistently lower performance on the WCB-L benchmark compared with their non-RP counterparts. Among all configurations, the \textit{non-RP + VGD + LCRL} paradigm achieves the strongest long-horizon planning capability. These findings indicate the differing perceptual demands of spatial grounding and long-horizon planning. Independent spatial grounding task focuses on mapping instructions to spatial coordinates, where RP acts as a strong reasoning regularizer that improves alignment. In contrast, long-horizon planning relies on hierarchical planning, and end-to-end optimization during mid-training can restrict generalization to more complex task structures. Despite these differences, both tasks consistently benefit from diverse augmented instruction–coordinate pairs, which improve data efficiency and reward density and serve as a task-agnostic form of \emph{knowledge enrichment} that broadly enhances model learning.

\subsubsection{Efficacy of CoT-SFT Mid-Training}
\label{sec:CoT-SFT_impact}

To evaluate the effectiveness of CoT-SFT mid-training, we conduct comparative experiments on Qwen2.5-VL-3B. Specifically, the process involves fine-tuning the model on 5k synthetic CoT samples until convergence. We design three controlled experimental settings (\textit{Directly LCRL}, \textit{+CoT-SFT}, and \textit{+SGRL+CoT-SFT}) to assess both the impact of CoT-SFT itself and its synergistic effect with spatial grounding mid-training.

\begin{table}[t]
\centering
\small
\setlength{\tabcolsep}{2pt} % 调整列间距，更紧凑
\begin{tabular}{l|c|cccc}
\hline
& GUI-R1-3B & Directly LCRL & +\textit{CoT-SFT} & +\textit{SGRL} & +Both \\
\hline
WCB-L & 0.487 & 0.603 & 0.629 & 0.632 & \textbf{0.658} \\
\hline
\end{tabular}
\caption{Comparative Analysis of Mid-Training and RL Integration Strategies on WCB-L}
\label{tab:wcbl4}
\end{table}

\begin{figure}[htbp]
     \centering
     \includegraphics[width=0.99\columnwidth]{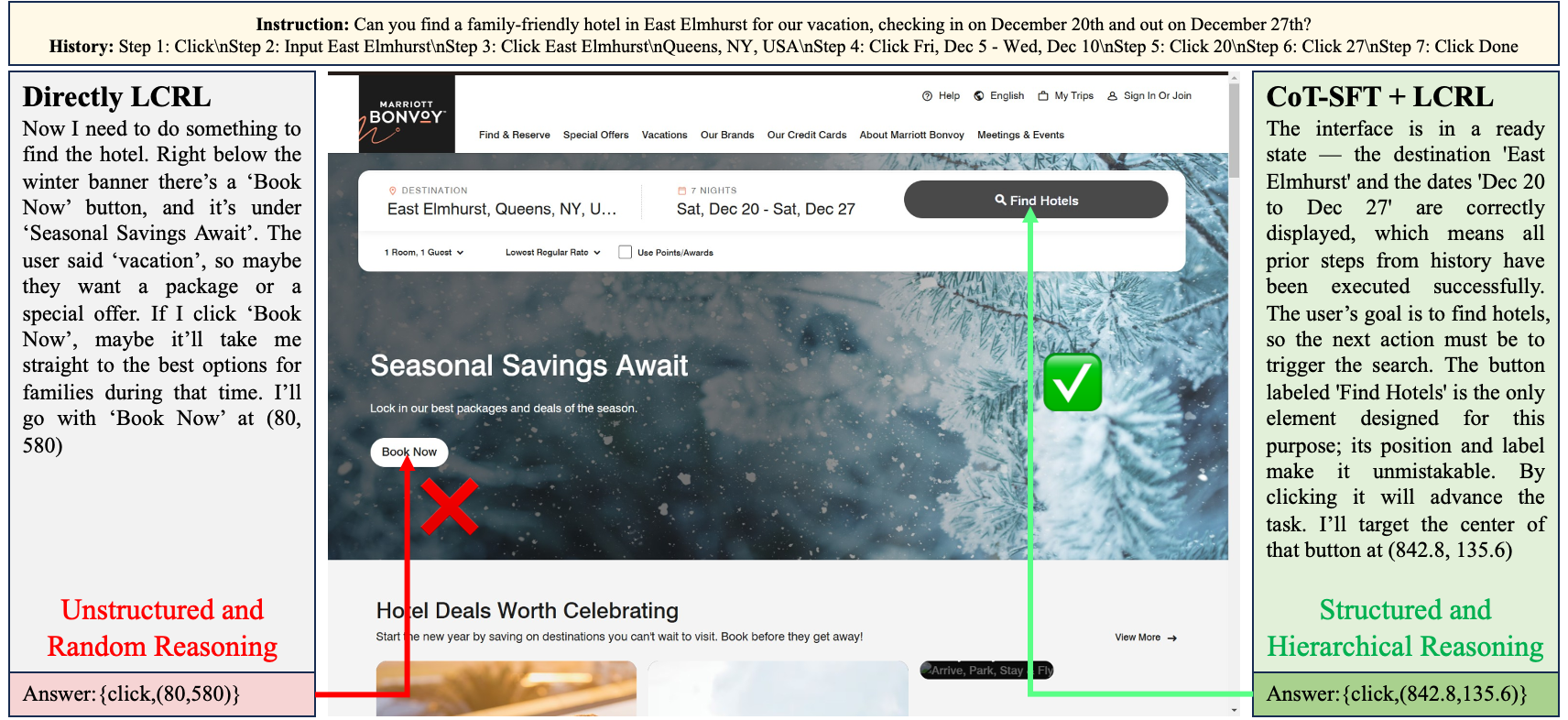} 
     \caption{Qualitative presentation of the imapact of CoT-SFT mid-training}
     \label{fig:cot_mid_impact}
 \end{figure}

\textbf{Analysis.} Quantitative results in Table~\ref{tab:wcbl4} show that CoT-SFT significantly improves downstream RL performance, and its combination with spatial grounding further elevates the model’s performance ceiling. Qualitative analyses in Figure~\ref{fig:cot_mid_impact} demonstrate that the mid-trained model produces longer, more structured reasoning traces, integrating visual observations, historical states, and high-level instructions. Together, the \textbf{Dual Mid-Training} paradigm disentangles spatial perception and temporal planning, allowing the final RL stage to focus on long-term reward optimization.

% -----------------------------------------------------------------------------
% 4.5 Public Benchmark
% -----------------------------------------------------------------------------
\subsection{Benchmark Evaluation and Comparison}

We evaluate our \dataset-trained models on a suite of GUI control benchmarks. The comparison in Table ~\ref{tab:wcbl4} and ~\ref{tab:public-benchmarks} indicates our \dataset-trained models consistently achieve strong performance across all benchmarks. 
For spatial grounding task, evaluated on AndroidControl-Low, GUI-Act-Web, and OmniAct-Desktop/Web, our models trained on WebChain with LCRL and augmented with CoT-SFT or SGRL achieve substantial improvements in both graphic rating and step success rate. For long-horizon planning task, evaluated on AndroidControl-High and GUI-Odyssey, which require multi-step planning and cross-app coordination, the same models show notable gains in task success rate compared to zero-shot or LCRL-only baselines. These results suggest that our \dataset-trained models demonstrate strong zero-shot and transfer performance across mobile, desktop, and web interfaces, underscoring the effectiveness of \dataset.

\section{Conclusion}
We have presented \dataset, a large-scale, open-source effort to standardize and accelerate research on web agents. By capturing human interactions on real-world websites and enforcing a \textit{Triple Alignment} between visual, structural, and action data, we provide the community with a robust foundation for building and evaluating models. Our experiments confirm the dataset's value, leading to the discovery of a superior \textit{Dual Mid-Training} strategy and setting a new performance benchmark on long-horizon tasks. The full release of our dataset, collection tools, and benchmark aims to break the reliance on proprietary data and foster transparent, reproducible GUI agent research.

\section*{Availability} The dataset, statistics, and usage instructions are released with this project. Detailed documentation is provided in the repository.

\section*{Ethics Statement}
We follow privacy-preserving data collection, redaction, and review. No personally identifiable information is released. We encourage responsible use and reporting of safety metrics.

{
    \small
    \bibliographystyle{ieeenat_fullname}
    \bibliography{main}
}


\end{document}